# Bridging Medical Data Inference to Achilles Tendon Rupture Rehabilitation

**An Qu*** **Cheng Zhang*** **Paul Ackermann†** **Hedvig Kjellström***

*Department of Robotics, Perception and Learning
KTH Royal Institute of Technology
Stockholm, Sweden
{anq, chengz, hedvig}@kth.se

†Department of Molecular Medicine and Surgery
Karolinska Institute
Stockholm, Sweden
paul.ackermann@karolinska.se

## Abstract

Imputing incomplete medical tests and predicting patient outcomes are crucial for guiding the decision making for therapy, such as after an Achilles Tendon Rupture (ATR). We formulate the problem of data imputation and prediction for ATR relevant medical measurements into a recommender system framework. By applying MatchBox, which is a collaborative filtering approach, on a real dataset collected from 374 ATR patients, we aim at offering personalized medical data imputation and prediction. In this work, we show the feasibility of this approach and discuss potential research directions by conducting initial qualitative evaluations.

## 1 Introduction

Despite advancements in treatment methods, recovery after musculoskeletal injuries, such as Achilles Tendon Rupture (ATR), is still a prolonged process with an unknown wide variation and often unsatisfactory outcome [1]. Current meta-analyses of the outcome after ATR have mainly focused on operative versus non-operative treatment, lately demonstrating no superiority of one over the other when early mobilization is applied [2]. However, when it comes to other factors and variables affecting the outcomes of ATR, there is currently a lack of knowledge. A recent study investigated how patient characteristics affected outcomes in male ATR patients. For example, it was found that increased age was a strong predictor of reduced function [3]. In an earlier work, factors such as pain and physical activities during rehabilitation were shown to be important for functional outcomes. However, the under-representation of female ATR patients has made it difficult to investigate the importance of gender on outcome [4]. Furthermore, the reason for the variability in outcome at least one year after ATR, however, still remains unknown. Hence, using machine learning tools to aid the study of ATR rehabilitation is desired.

Prediction of ATR rehabilitation is extremely challenging for both medical experts and machines due to numerous noisy measurements with large amount of missing data. Medical tests and outcome measurements for ATR, including the aforementioned ones, involve a large variety of metrics, and the total number of those metrics is on the magnitude of hundreds. Although many tests can be applied to patients to monitor the rehabilitation processes, the correlations between various measurements and patient characteristics are still under-explored. Thus, the test results are not fully informative to the clinicians. Moreover, some seemingly more reliable tests are expensive to perform, time consuming or uncomfortable for patients. Leveraged on the predictive power of data-driven approaches, it would be of great interest to find out whether we can use simple tests to predict possible results of those more reliable tests and potential outcomes for new patients.

Therefore, the task is to predict missing measurements and rehabilitation outcomes using existing noisy measurements. Since the rehabilitation out comes are a subset of measurements, we formulate the research question as data imputation thus we do not differ rehabilitation measures from other measurements. To this end, by formulating the data imputation and prediction of ATR relevant medical measurements into a recommender system framework, we apply MatchBox, which is a collaborative filtering approach [5], to 1) predict possible rehabilitation outcomes for patients; and 2) impute untested measurements for patients based on partially tested data. Our initial evaluation results show that the proposed method can predict a large set of medical measurements with high accuracy, while some measurements could not be well predicted due to reasons such as insufficient amount of training data, or weak medical relevance between the measurements and ATR rehabilitation etc.

## 2 Problem Statement

A dataset of ATR relevant medical measurements and rehabilitation outcomes have been collected at the Department of Orthopedics, Karolinska Institute Hospital, Solna, Sweden. The dataset composes medical records from 374 ATR patients involved in prospective randomized trials with 216 measurements, including routine medical tests, validated subjective outcome measurements and objective outcome measurements. Subjective outcomes are evaluated from the patient-reported questionnaires conducted at one year post operation, higher scores in those outcomes indicate better recovering. Objective outcomes are measured by ATR relevant medical examination conducted at one year post operation. Figure 1 illustrates the format of the employed dataset.

| | DVT_2w | inf_6w | ··· | Total_work_l | Height_max_l | ··· | ATRS_12m | FAOS_3_pain | ··· |
|---|---|---|---|---|---|---|---|---|---|
| patient_1 | 1 | X | ··· | X | 9.4 | ··· | 74 | X | ··· |
| patient_2 | 0 | 0 | ··· | 3811.7 | X | ··· | X | X | ··· |
| ⋮ | ⋮ | ⋮ | ⋮ | ⋮ | ⋮ | ⋮ | ⋮ | ⋮ | ⋮ |
| patient_n | X | X | ··· | 4425.0 | 13.1 | ··· | X | 97 | ··· |

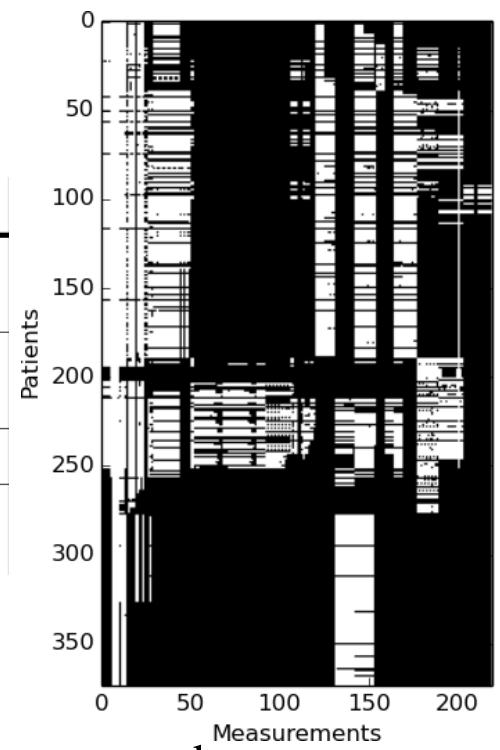

Figure 1: Left: a snapshot of the ATR relevant medical measurement dataset, each row represents a medical record of a patient including both routine tests and outcomes, and each column represents a measurement. The description for the entries is detailed in Appendix A. Right: an image showing the dataset by marking missing entries in black and the known entries in white.

In this dataset, however, 57205 measurement and outcome entries (as depicted in Figure 1), which account for 70% of the total number of entries in the dataset, are left untested due to health situations of patients, expensive medical cost or long time consumption required by the tests. Incomplete medical data record downgrades the effectiveness and accuracy for clinicians in analyzing patients' health status and rehabilitation progress, and then in turn affects making decisions for treatments.

Therefore, it is desired to enable the prediction of missing data, which can compensate for the unavailable information needed by clinicians, while not requiring extra resources to be committed by clinicians and patients. Referring to the dataset shown in Figure 1, this boils down to the widely addressed matrix completion problem. Methods such as mean imputation and last value carried forward have been investigated in simple scenarios [6]. When the matrix is of low rank, a convex optimization method was shown to produce exact result [7]. In cases of high rank matrices, [8] has shown good performance under the assumption that the columns of the matrix belong to a union of multiple low-rank subspaces. In the research line of recommender systems, Poisson factorization [9] and collaborative topic modeling [10, 11] have shown promising results, however, due to the underlying assumption of discrete rating space, they are not feasible for completing missing data that lies in continuous space.

In this work, we aim at offering personalized data prediction for patients, and the data amount is relatively small. Therefore, we employ a probabilistic graphical model based collaborative filtering framework, i.e. MatchBox [5] which is flexible on data type and does not require large amount of training data, for the imputation of missing tests and the prediction of rehabilitation outcomes. In this context, as both data imputation and prediction are conducted in the same framework, we will not distinguish between them in the presentation hereafter.

## 3 Methodology

Given a dataset of medical records with large amount of missing entries(Figure 1), we address the data prediciton problem using a recommender system framework that regards each patient as a user while regarding each of various medical records as an item rating. As such, the prediction of missing entries in the dataset is converted to the problem of predicting the ratings of items for a patient.

We adopt MatchBox [5], which is a generative model designed for personalized recommendation [5], for predicting the missing data in our dataset. Built upon collaborative filtering utilizing matrix factorization, MatchBox defines the low dimensional latent space as traits to encode latent properties of the data, and offers personalized predictions for individual users. Concretely, denoted by $A \in \mathbb{R}^{N\times M}$ the affinity matrix that represents patients' test results composed of $N$ patients and $M$ medical measurements, the prediction of an user $u's$ $i$-th medical test result is obtained by:

$$A_{u,i} = P_u R_i + B_u^P + B_i^R \tag{1}$$

where $P_u$ denotes the $u$-th row in the patient trait matrix $P \in \mathbb{R}^{N\times T}$ and $R_i$ is $i$-th column of the medical measurement trait matrix $R \in \mathbb{R}^{T\times M}$, $T$ indicates the number of traits selected. Additionally, $B_u^P$ is the $u$-th entry of the patient bias vector $B^P \in \mathbb{R}^N$, and $B_i^R$ represents the $i$-th entry of the medical measurement bias vector $B^R \in \mathbb{R}^M$. All elements in the aforementioned matrices are generated from independent Gaussian distributions, for which the means and variances need to be inferred. Thus, the priors are set as hyperparameters.

Note that MatchBox provides additionally content-based filtering to offer a hybrid system to encode user and item features for coping with also the cold-start problem. As an initial attempt in the ATR data prediction and given that we have many data points available for the majority of patients, we in this work adopt collaborative filtering only.

Importantly, our goal is to predict missing medical measurements that mostly lie in continuous space, which is different from the discrete rating system that thresholds the probabilistic outputs to produce a finite set of ratings. We thus adapted the implementation from Infer.NET [12] to model the measurement values directly, and omit the thresholding procedure to produce continuous outputs.

# 4 Evaluations

In this section, we evaluate the introduced problem formulation by applying collaborative filtering on the dataset collected from Achilles Tendon Rupture (ATR) patients. Prior to applying the prediction to the real dataset, we have tested the inference by latent parameter recovering using synthetic data as in [13]. The aim is to evaluate the inference performance since the latent variables are unknown to the model. The synthetic data was generated following the generative process of the model described in Section 3 with given priors. We were able to recover all the latent parameters in the model which indicate that the inference performs well.

As listed in Appendix A, the employed dataset composes ATR relevant medical test result for 216 measurements from 374 patients. Therefore, there are in total 80784 entries in the dataset, of which 57205 entries are missing.

Before inputting our dataset into the collaborative filter, we normalize the entry values to lie within $[0, 1]$ on a per measurement basis in terms of available data. Moreover, we have observed that the prior means and prior variances assigned to the patient trait matrix $P$, medical measurement trait matrix $R$, patient bias and medical measurement bias matrices $B^P$ and $B^R$ greatly affect the final prediction accuracy. As we do not have good knowledge for those priors, we fix the prior variances to be $0.5$, and then apply grid search for the prior means for each matrix. The grid search was conducted in the range of $[0.05, 0.95]$ with a fixed step size of $0.05$.

**Results** For the experiment, we conducted a 5-fold cross validation for the whole dataset composed by 374 patients and 216 measurements. For evaluating the performance of our approach, in the 5-fold cross validation we predict the measurement values of every test set, which we have the true values of, and compare the results against the ground truth. The mean errors of our predictions are reported in Figure 2a. We can see that almost half of the measurements can be predicted with mean errors smaller than $0.2$, which is relatively small to be able to indicate the real test results. Furthermore, we also observe that the majority of inaccurate predictions were associated with the measurements which have very little amount of known data. This is expected since the model was not able to generalize over different values for those measurements with insufficient training data.

In order to evaluate the method without being affected by the insufficiency of data, we conducted the same experiments for the $125$ measurements, each of which has more than $50$ known data for training and testing. As reported in Figure 2b, we can see an enhancement in prediction mean errors

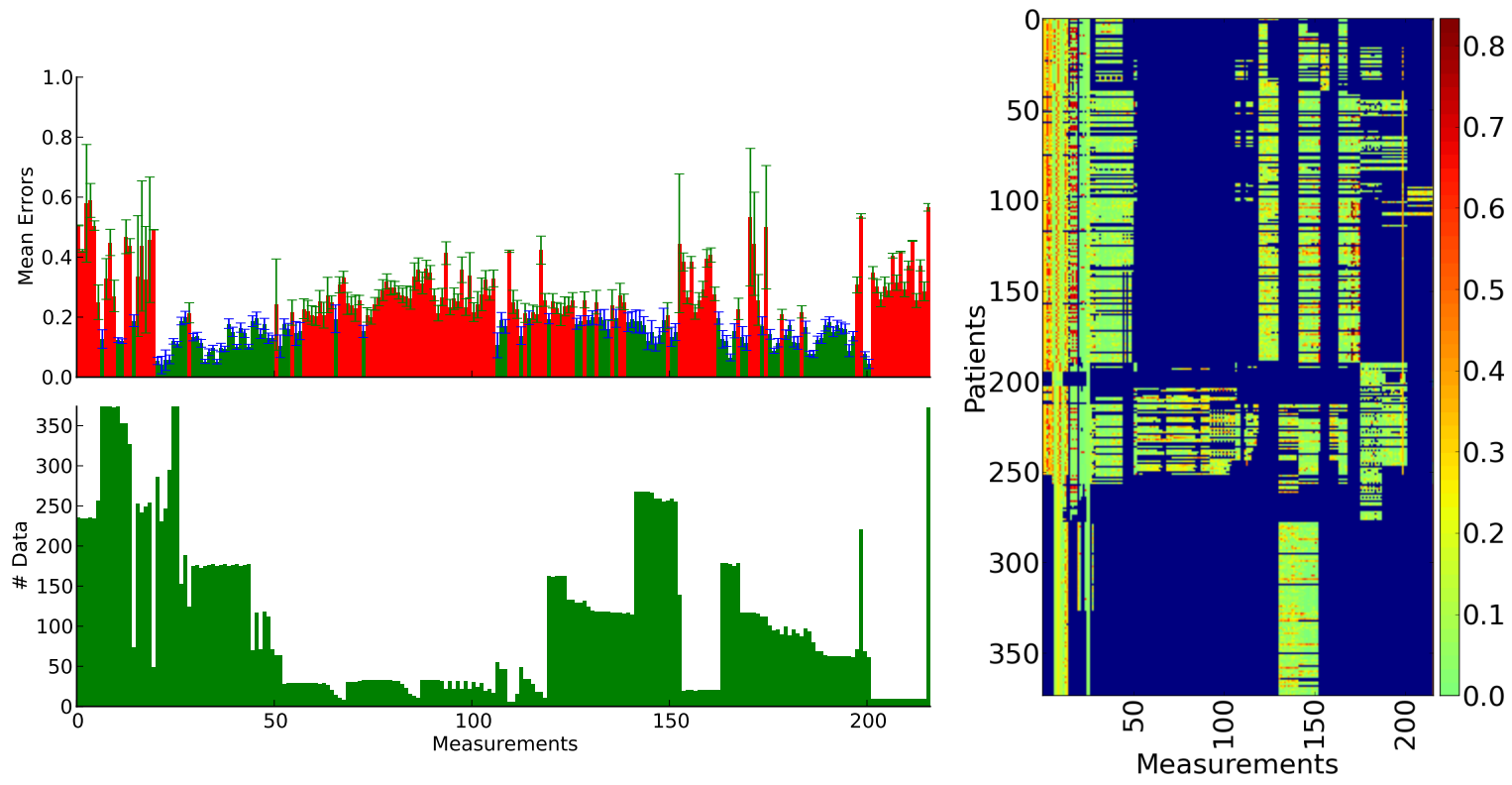

(a) Results for 216 measurements. Mean error over all measurements: $0.2478$.

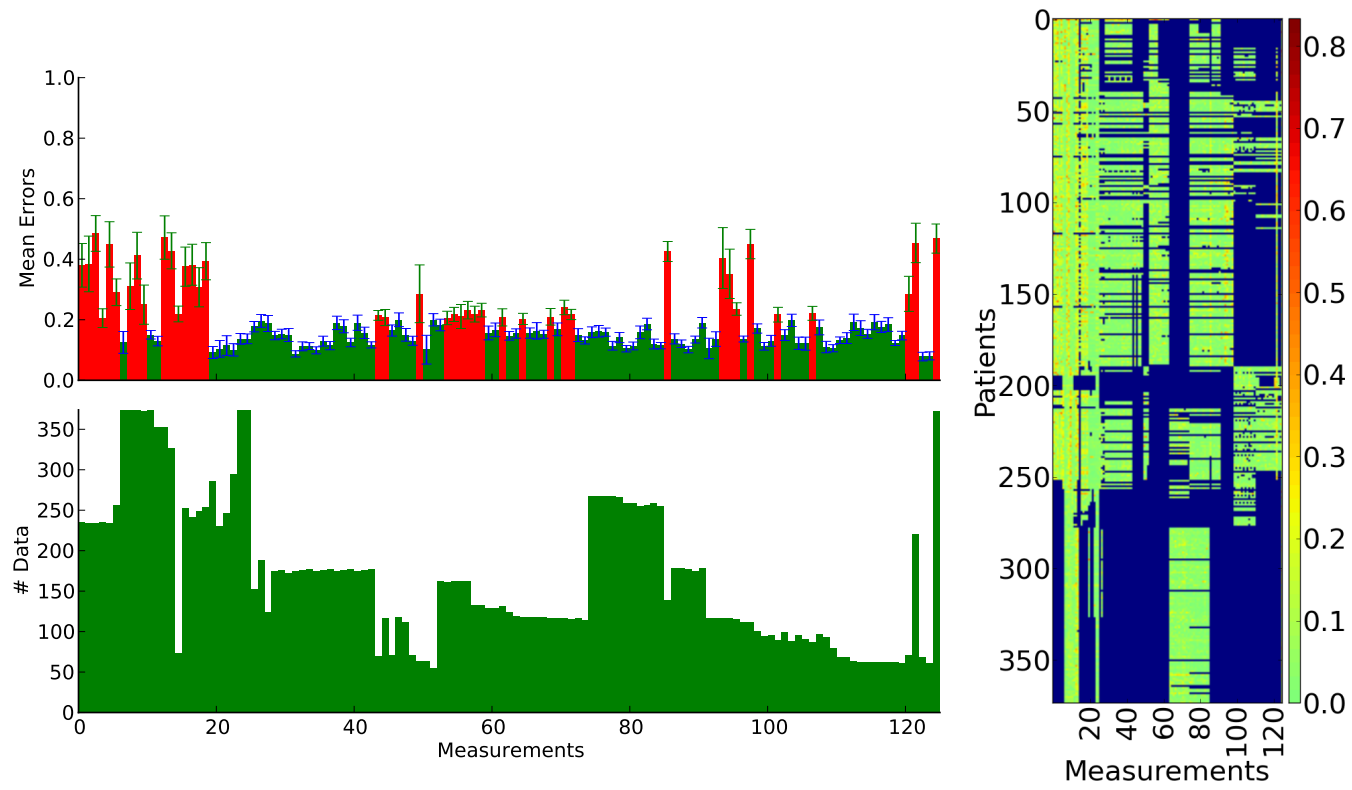

(b) Results for 125 measurements. Mean error over all measurements: $0.2010$.

Figure 2: In both (a) and (b), left side shows prediction mean errors for each measurement (upper), we color the measurement in green if the prediction for it was relatively small (mean error $\leq 0.2$) and in red otherwise. The number of known data points per measurement (lower) is also shown. The right side shows the prediction mean errors for each data point with colors indicating the magnitudes, the blue areas are the missing data points without available ground truth.

which decreased from $0.2478$ to $0.2010$. However, it is worthwhile to note that there are still some measurements (red bars) that can not be well predicted, although the amount of known data for them is large. By taking a closer look at the data, as listed in Appendix A, we have observed that those inaccurate predictions were made mostly for either medically irrelevant measurements, e.g., date of test, or the measurements that have binary outputs, which were not well predicted as our framework models measurements in continuous space and is not capable of handling those measurements.

# 5 Conclusion

Because some of the medical tests of interest are expensive to perform, time consuming or uncomfortable for patients, the amount of missing data is unavoidably large. As desired by clinicians, more medical test results will be helpful in medical diagnosis and making decisions for treatments. Therefore, we in this work investigated the possibility of formulating the medical data prediction problem into a recommender system framework. By modeling the patients as users and medical test results as item ratings, we show that it is feasible to predict missing medical data using collaborative filtering, and the evaluation results indicated that we could achieve good prediction accuracy for many medical measurements.

As a potential future work, we would like to separate the measurement data types, e.g., continuous or discrete values, to decide whether a thresholding is needed for output to improve the prediction accuracy. Furthermore, rather than pure collaborative filtering, it would be interesting to adopt the hybrid system provided by MatchBox to investigate further improvements. For better evaluation of the performance, we plan to compare our method with other matrix completion approaches, as well as tracking the rehabilitation results of patients to justify the usefulness of our method.

# Appendix A List of Medical Measurements in the Achilles Tendon Rupture Patients' Dataset

| | Measurements | Description |
|---|---|---|
| 1 | Complication | Signs in journal documentation of any complication? |
| 2 | Paratenon | Closed the paratenon? |
| 3 | Fascia | Closed the fasica cruris? |
| 4 | PDS | Used 2 PDS sutures? |
| 5 | Surg_comp | Surgeon catogorized as compliant to op-schedule? |
| 6 | Treatment_Group | Treatment Group For all studies |
| 7 | Plast_foot | A - Plastercast + Foot-IPC |
| 8 | Ort_B | B - Calf IPC |
| 9 | Plast | A, B, D plastercast |
| 10 | Vacoped | B, C, D - Vacoped |
| 11 | Time_to_op | Time to Operation |
| 12 | Op_time | Operation Time |
| 13 | Op_time_dic | Dicomisation of op. time |
| 14 | EXP | Specialist Y/N |
| 15 | Compliant | Compliant |
| 16 | DVT_2 | Deep venous thrombosis after two weeks? |
| 17 | DVT_6w | DVT after 6 weeks? |
| 18 | DVT_2w_and_6w | DVT at both 2 and 6 weeks? |
| 19 | DVT_2w_or_6w | Total DVT after 2 and 6 weeks? |
| 20 | DVT_8w | Only for GBG Studies |
| 21 | Inf_2w | Infection at 2 weeks post.op? |
| 22 | Inf_6w | Infection at 6 weeks post.op? |
| 23 | Any_inf | Infection at either 2 or 6 weeks postop? |
| 24 | Rerupture | Rerupture? |
| 25 | Pump_pat_reg | Patient registered pump time (hh:mm:ss) |
| 26 | Pump_reg | Device registered pump time (hh:mm:ss) |
| 27 | Preinjury | Degree of activity preinjury |
| 28 | Post_op | Degree of activity one year post operation |
| 29 | D_PAS | Delta PAS = Pre-Post |
| 30 | Con_Power_I | Concentric power (Watt) - Injured side |
| 31 | Con_Power_U | Concentric power (Watt) - Uninjured side |
| 32 | LSI_Con_Power | Limb Symmetry Index - Concentric power |
| 33 | Total_work_I | Total concentric work (Joule) - Injured side |
| 34 | Total_work_U | Total concentric work (Joule) - Uninjured side |
| 35 | LSI_Total_work | Limb Symmetry Index - Total work |
| 36 | Repetition_I | Number of heel-rises - Injured side |
| 37 | Repetition_U | Number of heel-rises - Uninjured side |
| 38 | LSI_Repetitions | Limb Symmetry Index - Repetitions |
| 39 | Height_Max_I | Maximum height of heel-rise (Centimeter) - Injured side |
| 40 | Height_Max_U | Maximum height of heel-rise (Centimeter) - Uninjured side |
| 41 | LSI_Height | Limb Symmetry Index - Height |
| 42 | Height_A_I | Average height of heel-rise (Centimeter) - Injured side |
| 43 | Height_A_U | Average height of heel-rise (Centimeter) - Uninjured side |
| 44 | LSI_Height_Ave | Limb Symmetry Index ? Height_Average |
| 45 | Ecc_Power_I | Eccentric power (Watt) - Injured side |
| 46 | Height_Min_I | Mimimum height of heel-rise (Centimeter) - Injured side |
| 47 | Ecc_Power_U | Eccentric power (Watt) - Uninjured side |
| 48 | Height_Min_U | Mimimum height of heel-rise (Centimeter) - Uninjured side |
| 49 | LSI_Height_Min | Limb Symmetry Index - Height_Min |
| 50 | LSI_Ecc_Power | Limb Symmetry Index - Eccentric power |
| 51 | Muscle_vein_thrombosis_2 | Muscle vein thrombosis 2 weeks follow-up |
| 52 | Thompson_2 | Thompson 2 weeks follow-up |
| 53 | Podometer_day1 | Podometer day 1 |
| 54 | Podometer_day2 | Podometer day 2 |
| 55 | Podometer_day3 | Podometer day 3 |
| 56 | Podometer_day4 | Podometer day 4 |
| 57 | Podometer_day5 | Podometer day 5 |
| 58 | Podometer_day6 | Podometer day 6 |
| 59 | Podometer_day7 | Podometer day 7 |
| 60 | Podometer_day8 | Podometer day 8 |
| 61 | Podometer_day9 | Podometer day 9 |
| 62 | Podometer_day10 | Podometer day 10 |
| 63 | Podometer_day11 | Podometer day 11 |
| 64 | Podometer_day12 | Podometer day 12 |
| 65 | Podometer_day13 | Podometer day 13 |
| 66 | Podometer_day14 | Podometer day 14 |
| 67 | Podometer_day15 | Podometer day 15 |
| 68 | Podometer_day16 | Podometer day 16 |
| 69 | Podometer_on_day_of_microdialysis | Podometer on day of microdialysis |
| 70 | Podometer_on_day_minus_1 | Podometer on day before microdialysis |
| 71 | Podometer_on_day_minus_2 | Podometer two days before microdialysis |
| 72 | Subjective_load_day1 | Subjective load day 1 |
| 73 | Subjective_load_day2 | Subjective load day 2 |
| 74 | Subjective_load_day3 | Subjective load day 3 |
| 75 | Subjective_load_day4 | Subjective load day 4 |
| 76 | Subjective_load_day5 | Subjective load day 5 |
| 77 | Subjective_load_day6 | Subjective load day 6 |
| 78 | Subjective_load_day7 | Subjective load day 7 |
| 79 | Subjective_load_day8 | Subjective load day 8 |
| 80 | Subjective_load_day9 | Subjective load day 9 |
| 81 | Subjective_load_day10 | Subjective load day 10 |
| 82 | Subjective_load_day11 | Subjective load day 11 |
| 83 | Subjective_load_day12 | Subjective load day 12 |
| 84 | Subjective_load_day13 | Subjective load day 13 |
| 85 | Subjective_load_day14 | Subjective load day 14 |
| 86 | Subjective_load_day15 | Subjective load day 15 |
| 87 | Subjective_load_day16 | Subjective load day 16 |
| 88 | Days_until_microdialysis | Days postsurgery until microdialysis |
| 89 | Load_on_day_of_microdialysis | LOAD |
| 90 | Load_on_day_minus_1 | LOAD_Minus1 |
| 91 | Load_on_day_minus_2 | LOAS_Minus2 |
| 92 | Number_of_days_with_load | Load_Days |
| 93 | VAS_day1_act | VAS day 1, active |
| 94 | VAS_day1_pas | VAS day 1, passive |
| 95 | VAS_day2_act | VAS day 2, active |
| 96 | VAS_day2_pas | VAS day 2, passive |
| 97 | VAS_day3_act | VAS day 3, active |
| 98 | VAS_day3_pas | VAS day 3, passive |
| 99 | VAS_day4_act | VAS day 4, active |
| 100 | VAS_day4_pas | VAS day 4, passive |
| 101 | VAS_day5_act | VAS day 5, active |
| 102 | VAS_day5_pas | VAS day 5, passive |
| 103 | VAS_day6_act | VAS day 6, active |
| 104 | VAS_day6_pas | VAS day 6, passive |
| 105 | VAS_day7_act | VAS day 7, active |
| 106 | VAS_day7_pas | VAS day 7, passive |
| 107 | VAS_injured_2weeks | VAS injured, 2 weeks |
| 108 | Calf_circumference_injured_mean | Calf circumference -injured tendon, mean value, week 2 |
| 109 | Calf_circumference_control_mean | Calf circumference -control tendon, mean value, week 2 |
| 110 | Plantar_flexion_injured_mean | Plantar flexion -injured tendon, mean value, week 2 |
| 111 | Plantar_flexion_control_mean | Plantar flexion -control tendon, mean value, week 2 |
| 112 | Dorsal_flexion_injured_mean | Dorsal flexion -injured tendon, mean value, week 2 |
| 113 | Dorsal_flexion_control_mean | Dorsal flexion -control tendon, mean value, week 2 |
| 114 | Heel_rise_average_height_injured_6mo | heel raise injured 6 months |
| 115 | Heel_rise_average_height_control_6mo | heel raise uninjured 6 months |
| 116 | difference_heel_raise_6mo | difference heel raise 6 months |
| 117 | Heel_rise_average_height_injured_1yr | heel raise injured 1 year |
| 118 | Heel_rise_average_height_control_1yr | heel raise uninjured 1 year |
| 119 | difference_heel_raise_1yr | difference heel raise 1 year |
| 120 | ATRS_3_Strenght | Achilles Tendon Rupture score 3 months - Limited strength in the calf/tendon/foot? |
| 121 | ATRS_3_tired | Achilles Tendon Rupture score 3 months - Tired in the calf/tendon/foot? |
| 122 | ATRS_3_stiff | Achilles Tendon Rupture score 3 months - Stiffness in the calf/tendon/foot? |
| 123 | ATRS_3_pain | Achilles Tendon Rupture score 3 months - Pain in the calf/tendon/foot? |
| 124 | ATRS_3_ADL | Achilles Tendon Rupture score 3 months - limited ADL? |
| 125 | ATRS_3_Surface | Achilles Tendon Rupture score 3 months - limited on uneven surface? |
| 126 | ATRS_3_stairs | Achilles Tendon Rupture score 3 months - limited in stairs/hills? |
| 127 | ATRS_3_run | Achilles Tendon Rupture score 3 months - limited when running? |
| 128 | ATRS_3_jump | Achilles Tendon Rupture score 3 months - limited when jumping? |
| 129 | ATRS_3_phys | Achilles Tendon Rupture score 3 months - limited in physical work? |
| 130 | ATRS_3_Sum | Achilles Tendon Rupture score 3 months - total sum |
| 131 | ATRS_item1_6month | ATRS 6month - Limited strength in the calf/tendon/foot? |
| 132 | ATRS_item2_6month | ATRS 6month - Tired in the calf/tendon/foot? |
| 133 | ATRS_item3_6month | ATRS 6month - Stiffness in the calf/tendon/foot? |
| 134 | ATRS_item4_6month | ATRS 6month - Pain in the calf/tendon/foot? |
| 135 | ATRS_item5_6month | ATRS 6month - limited ADL? |
| 136 | ATRS_item6_6month | ATRS 6month - limited on uneven surface? |
| 137 | ATRS_item7_6month | ATRS 6month - limited in stairs/hills? |
| 138 | ATRS_item8_6month | ATRS 6month - limited when running? |
| 139 | ATRS_item9_6month | ATRS 6month - limited when jumping? |
| 140 | ATRS_item10_6month | ATRS 6month - limited in physical work |
| 141 | ATRS_total_score_6month | ATRS total score 6month |
| 142 | ATRS_12_strength | Achilles Tendon Rupture score 12 months - Limited strength in the calf/tendon/foot? |
| 143 | ATRS_12_tired | Achilles Tendon Rupture score 12 months - Tired in the calf/tendon/foot? |
| 144 | ATRS_12_stiff | Achilles Tendon Rupture score 12 months - Stiffness in the calf/tendon/foot? |
| 145 | ATRS_12_pain | Achilles Tendon Rupture score 12 months - Pain in the calf/tendon/foot? |
| 146 | ATRS_12_ADL | Achilles Tendon Rupture score 12 months - limited ADL? |
| 147 | ATRS_12_Surface | Achilles Tendon Rupture score 12 months - limited on uneven surface? |
| 148 | ATRS_12_stairs | Achilles Tendon Rupture score 12 months - limited in stairs/hills? |
| 149 | ATRS_12_run | Achilles Tendon Rupture score 12 months - limited when running? |
| 150 | ATRS_12_jump | Achilles Tendon Rupture score 12 months - limited when jumping? |
| 151 | ATRS_12_phys | Achilles Tendon Rupture score 12 months - limited in physical work? |
| 152 | ATRS_12m | Achilles Tendon Rupture Score, 12 month control |
| 153 | ATRS_2cl | ATRS> 80.0? |
| 154 | FAOS_3_Pain | FAOS Pain 3 months |
| 155 | FAOS_3_Symptom | FAOS Symptom 3 months |
| 156 | FAOS_3_ADL | FAOS ADL 3 month |
| 157 | FAOS_3_sport_rec | FAOS sport and recreation 3 months |
| 158 | FAOS_3_QOL | FAOS QOL 3 months |
| 159 | FAOS_6_Symptom | FAOS Symptom 6month |
| 160 | FAOS_6_Pain | FAOS Pain 6month |
| 161 | FAOS_6_ADL | FAOS ADL 6month |
| 162 | FAOS_6_Sport_Rec | FAOS Function 6month |
| 163 | FAOS_6_QOL | FAOS QOL 6month |
| 164 | FAOS_12_Pain | FAOS 1 year Pain - 100=No pain |
| 165 | FAOS_12_Symptom | FAOS 1 year Symptom - 100=No symtoms |
| 166 | FAOS_12_ADL | FAOS 1 year Activity of daily living - 100=Full ADL |
| 167 | FAOS_12_Sport_Rec | FAOS 1 year Function in sport and recreation - 100=No problems |
| 168 | FAOS_12_QOL | FAOS 1 year Foot and ankle-related Quality of Life - 100=full QoL |
| 169 | Q1 | Q1(mobility) |
| 170 | Q3 | Q3(activities) |
| 171 | Q4 | Q4(pain) |
| 172 | Q5 | Q5(depression) |
| 173 | EQ5D_ix | EQ5D index |
| 174 | VAS | EuroQoL Visual Analogue Scale |
| 175 | VAS_2 | Dichotomized EQ VAS |
| 176 | GLUC_injured_mean | Glucose - Injured tendon, mean value, week 2 |
| 177 | GLUC_control_mean | Glucose - Control tendon, mean value, week 2 |
| 178 | LAC_injured_mean | Lactate - Injured tendon, mean value, week 2 |
| 179 | LAC_control_mean | Lactate - Control tendon, mean value, week 2 |
| 180 | PYR_injured_mean | Pyruvate - Injured tendon, mean value, week 2 |
| 181 | PYR_control_mean | Pyruvate - Control tendon, mean value, week 2 |
| 182 | GLY_injured_mean | Glycerol - Injured tendon, mean value, week 2 |
| 183 | GLY_control_mean | Glycerol - Control tendon, mean value, week 2 |
| 184 | GLUT_injured_mean | Glutamate - Injured tendon, mean value, week 2 |
| 185 | GLUT_control_mean | Glutamate - Control tendon, mean value, week 2 |
| 186 | LAC2_PYR2_ratio_injured_mean | Lactate-Pyruvate ratio -Injured tendon, mean value, week 2 |
| 187 | LAC2_PYR2_ratio_control_mean | Lactate-Pyruvate ratio -Control tendon, mean value, week 2 |
| 188 | PINP_injured | Collagen I marker (pg/ml) - injured leg |
| 189 | PIIINP_injured | Collagen III marker (pg/ml) - injured leg |
| 190 | Bradford_injured | Protein concentration in injured leg (ug/ml) |
| 191 | PINP_normalized_Injured | Collagen I marker normalized to protein conc. (pg/ul) - injured leg |
| 192 | PIIINP_normalized_injured | Collagen III marker normalized to protein conc. (pg/ul) - injured leg |
| 193 | PINP_uninjured | Collagen I marker (pg/ml) - uninjured leg |
| 194 | PIIINP_uninjured | Collagen III marker (pg/ml) - uninjured leg |
| 195 | Bradford_uninjured | Protein concentration in uninjured leg (ug/ml) |
| 196 | PINP_normalized_Uninjured | Collagen I marker normalized to protein conc. (pg/ul) - uninjured leg |
| 197 | PIIINP_normalized_uninjured | Collagen III marker normalized to protein conc. (pg/ul) - uninjured leg |
| 198 | Collagen | Collagen Analysis |
| 199 | Glut_2_inj_values | Complete Glut_2_inj values? |
| 200 | P_ratio_inj | PINP/PIIINP ratio injured |
| 201 | P_ratio_uninj | PINP/PIIINP ratio uninjured leg |
| 202 | RF_injured | INJURED - Resting flux: the median flux value obtained during baseline |
| 203 | RF_uninjured | uninjured - Resting flux: the median flux value obtained during baseline |
| 204 | BZ_injured | INJURED - Biological zero: the median flux value obtained during occlusion |
| 205 | BZ_uninjured | uninjured - Biological zero: the median flux value obtained during occlusion |
| 206 | MF_injured | INJURED - Maximum flux: the highest flux value obtained during PORH |
| 207 | MF_uninjured | uninjured - Maximum flux: the highest flux value obtained during PORH |
| 208 | T_RF_injured | INJURED - Time to resting flux. the time from t0 until RF is reached |
| 209 | T_RF_uninjured | uninjured - Time to resting flux. the time from t0 until RF is reached |
| 210 | T_MF_injured | INJURED - Time to maximum flux: the time from t0 until MF is reached |
| 211 | T_MF_uninjured | uninjured - Time to maximum flux: the time from t0 until MF is reached |
| 212 | T_HR_injured | INJURED - Time to half recovery: the time from t0 after tMF until MF + RF is reached |
| 213 | T_HR_uninjured | uninjured - Time to half recovery: the time from t0 after tMF until MF + RF is reached |
| 214 | Ratio_MF_RF_injured | INJURED - Ratio of maximum flux and resting flux |
| 215 | Ratio_MF_RF_uninjured | uninjured - Ratio of maximum flux and resting flux |
| 216 | B1_D66 | B1-150, C3,4,6,7,17-25, D2-D66 |